 \documentclass[pmlr,twocolumn,10pt]{jmlr} 

\usepackage{booktabs}
\usepackage{siunitx}

\usepackage[switch]{lineno}

\usepackage{cleveref}

\theorembodyfont{\upshape}
\theoremheaderfont{\scshape}
\theorempostheader{:}
\theoremsep{\newline}

\jmlrvolume{LEAVE UNSET}
\jmlryear{2024}
\jmlrsubmitted{LEAVE UNSET}
\jmlrpublished{LEAVE UNSET}
\jmlrworkshop{Machine Learning for Health (ML4H) 2024} 

\title[Weakly-Supervised Multimodal Learning on MIMIC-CXR]{Weakly-Supervised Multimodal Learning on MIMIC-CXR}

\author{ 
  \Name{Andrea Agostini}$^1$\thanks{Corresponding author: \texttt{andrea.agostini@inf.ethz.ch}}, \Name{Daphné Chopard}$^{1,2}$, \Name{Yang Meng}$^{3}$,  
  \Name{Norbert Fortin}$^4$, \\ \Name{Bahbak Shahbaba}$^3$, \Name{Stephan Mandt}$^{3,5}$, \Name{Thomas {M.~Sutter}}$^{1,\dagger}$, \Name{Julia {E. Vogt}}$^{1,\dagger}$ \\
  \\
    \addr
    $^1$Department of Computer Science, ETH Zurich \\
    $^2$Department of Intensive Care and Neonatology, University Children’s Hospital Zurich \\
    $^3$Department of Statistics, UC Irvine \\
    $^4$Department of Neurobiology and Behavior, UC Irvine \\
    $^5$Department of Computer Science, UC Irvine \\
    $^\dagger$ Shared last Authorship
}

\usepackage{caption}
\begin{document}

\maketitle
\begin{abstract}
Multimodal data integration and label scarcity pose significant challenges for machine learning in medical settings.
To address these issues, we conduct an in-depth evaluation of the newly proposed Multimodal Variational Mixture-of-Experts (MMVM) VAE on the challenging MIMIC-CXR dataset.
Our analysis demonstrates that the MMVM VAE consistently outperforms other multimodal VAEs and fully supervised approaches, highlighting its strong potential for real-world medical applications.


\end{abstract}

\paragraph*{Data and Code Availability}
In this work, we use the JPG version of the MIMIC-CXR Database, named MIMIC-CXR-JPG \citep{johnson2019mimic,  johnson2024mimic}. The \hyperlink{https://physionet.org/content/mimic-cxr-jpg/2.1.0/}{dataset}\footnote{\scriptsize \url{https://physionet.org/content/mimic-cxr-jpg/2.1.0/}} and the \hyperlink{https://github.com/agostini335/mmvmvae-mimic}{code}\footnote{\scriptsize \url{https://github.com/agostini335/mmvmvae-mimic}} used to run our experiments are publicly available.

\paragraph*{Institutional Review Board (IRB)}
This work does not require IRB approval.


\section{Introduction}

Medical data is inherently multimodal, encompassing a variety of data types such as electronic health records, medical images, omics, and clinical notes.
Integrating these different data modalities is crucial in modern medical practice, particularly in fields like radiology and precision oncology. By integrating information from multiple sources, clinicians can develop a more comprehensive understanding of patients, leading to more accurate diagnoses and improved treatments \citep{zhou2024multimodal,huang2020fusion}.
However, effectively combining and jointly analyzing data from such a diverse set of modalities presents a significant challenge for both medical practitioners and current machine learning (ML) models.

Despite the demonstrated potential of ML to enhance clinical decision-making, the complexity of the medical domain has limited its widespread application \citep{miotto2018deep}.
One major barrier is the scarcity of high-quality annotated datasets, which are costly to produce due to the need for expert healthcare professionals \citep{hassanzadeh2018clinical}.
These challenges underscore the need for ML approaches that can learn from a limited number of labeled examples, effectively leverage multiple modalities, and handle missing modalities. 

Multimodal Variational Autoencoders (VAEs) have emerged as a promising solution for jointly modeling and learning from weakly-supervised heterogeneous data sources by learning a joint posterior approximation distribution \citep{wu_multimodal_2018,shi_variational_2019,sutter2021}.
While multimodal VAEs learning a joint posterior approximation distribution are scalable and capable of effectively managing multiple modalities, trying to fuse all the multimodal information into a shared latent space may lead to insufficient latent representations \citep{daunhawer2021limitations}.
To address these limitations, the Multimodal Variational Mixture-of-Experts prior (MMVM) VAE \citep{sutter2024} introduces a soft-sharing mechanism that allows each modality to contribute more flexibly to a shared aggregate posterior. This approach results in superior latent representations and enables each encoding to better preserve information from its original, uncompressed features.

In this paper, we present a comprehensive evaluation of the MMVM VAE method focused on its applicability in medical settings. Our experiments, conducted on the challenging MIMIC-CXR radiology dataset, focus on the task of diagnostic label prediction. First, we benchmark the learned latent representations of the MMVM VAE against other state-of-the-art multimodal VAEs. Second, we compare the performances of the MMVM VAE and fully supervised approaches regarding their sensitivity to label availability. This evaluation provides valuable insights into the critical challenges associated with multimodal healthcare data and demonstrates the method's ability to effectively address these challenges, showcasing its potential for applied settings.

\vspace{-0.2cm}
\section{Methods}
\label{sec:methods}

We consider a dataset $\mathbb{X} = \{ \Vec{X}^{(i)} \}_{i=1}^n$ where each $\Vec{X}^{(i)} = \{ \vec{x}_1^{(i)}, \ldots, \vec{x}_M^{(i)} \}$ is a set of $M$ modalities $\vec{x}_m$ with latent variables $\vec{z} = \{ \vec{z}_1^{(i)}, \ldots, \vec{z}_M^{(i)} \}$.

Commonly, multimodal VAEs \citep[e.g., ][]{wu_multimodal_2018,shi_variational_2019,sutter2021} optimize the following ELBO objective:
\begin{align}
    \mathcal{E} (\vec{X}) = &~ \mathbb{E}_{q_\phi( \vec{z} \mid \vec{X})} \left[ \log \frac{p_\theta (\vec{X} \mid \vec{z})}{q_\phi( \vec{z} \mid \vec{X})} + \log p( \vec{z} ) \right]. \nonumber
\end{align}
This objective involves a decoder (or likelihood) $p_\theta (\vec{X} \mid \vec{z})$ and an encoder (or variational distribution) ${q_\phi(\vec{z} \mid \vec{X})}$, where $\theta$ and $\phi$ denote the learnable model variational parameters.

To efficiently handle missing modalities during inference, the joint variational distribution takes the form \citep{sutter2021}
\begin{align}
    q_\phi(\vec{z} \mid \vec{X}) = f_{agg} (q_\phi (\vec{z} \mid \vec{x}_1), \ldots, q_\phi (\vec{z} \mid \vec{x}_M)), \nonumber
\end{align}
where $f_{agg}(\cdot)$ defines any aggregation function that combines the unimodal posterior approximations $q_\phi (\vec{z} \mid \vec{x}_m)$ into the joint posterior approximation.
Throughout this work, we refer to multimodal VAEs that follow this principle as aggregation-based multimodal VAEs \citep[e.g., ][]{wu_multimodal_2018, shi_variational_2019, sutter2021}.

In contrast to aggregation-based multimodal VAEs, which rely on learning a joint posterior approximation function, the MMVM VAE leverages a data-dependent prior ${h( \vec{z} \mid \vec{X} )}$ that enables the \emph{soft-sharing} of information between different modalities \citep{sutter2024}.

The prior ${h( \vec{z} \mid \vec{X} )}$, which models the dependency between the different modalities, is defined as
\begin{align*}
    h(\vec{z} \mid \vec{X} ) = &~  \prod_{m=1}^M h(\vec{z}_m \mid \vec{X}),~ \text{where} \nonumber \\
    h(\vec{z}_m \mid \vec{X}) = &~  \frac{1}{M} \sum_{\tilde{m}=1}^M q_\phi (\vec{z}_m \mid \vec{x}_{\tilde{m}}), \forall m \in M.
\vspace{-0.2cm}
\end{align*}
More specifically, the MMVM VAE optimizes the following ELBO-like learning objective $\mathcal{E}$
\begin{align*}
    \mathcal{E} (\vec{X}) = &~ \sum_{m=1}^M \mathbb{E}_{q_\phi( \vec{z}_m \mid \vec{x}_m)} \left[ \log \frac{p_\theta (\vec{x}_m \mid \vec{z}_m)}{q_\phi( \vec{z}_m \mid \vec{x}_m)} \right]  \nonumber \\
    &~ + \mathbb{E}_{q_\phi( \vec{z} \mid \vec{X})} \left[ {\log h( \vec{z} | \vec{X} )} \right].
\vspace{-0.2cm}
\end{align*}

This objective encourages similarity between the unimodal posterior approximations, facilitating a more cohesive latent representation across modalities.

\section{The MIMIC-CXR Dataset}   
Experiments are conducted on the MIMIC-CXR dataset, a well-established and large collection of chest X-rays. The dataset reflects real clinical challenges with varying image quality due to technical issues, patient positioning, and obstructions. Chest X-rays are captured from various view positions, which provide valuable complementary information that can be used to improve diagnostics \citep{raoof2012interpretation}. The dataset is organized on a study level, where each study is defined as a collection of images linked to a single radiology report. Studies are annotated with fourteen diagnostic labels, automatically extracted from the associated report.

In this work, we consider \textit{frontal} and \textit{lateral} view positions as two distinct modalities, i.e., \mbox{$\vec{X} = \{ \vec{x}_f, \vec{x}_l  \}$}.
Further details on data preprocessing, dataset splitting, and examples of these bimodal tuples are provided in Appendix~\ref{app:sec:Dataset}.
\vspace{-0.2cm}
\section{Experiments}
We assess the MMVM VAE's ability to learn meaningful representations from multimodal data, addressing the task of diagnostic label prediction. We evaluate it against other multimodal VAEs and investigate the effect of limited label availability, by comparing it to fully-supervised approaches.
\vspace{-0.2cm}

\subsection{Comparison of Multimodal VAEs}
\label{sec:latent_representations_eval}
In this experiment, we first train multimodal VAEs, then use the learned latent representations to solve the classification task.



\paragraph{Baselines}
We compare the MMVM VAE to different aggregation-based multimodal VAEs (see \cref{sec:methods}) and a set of independent VAEs \citep[independent, ][]{kingma2013}, jointly trained on both modalities.
In the independent VAEs, each modality is trained separately, without interaction or regularization across modalities. 
In addition, we evaluate four aggregation-based multimodal VAEs with different aggregation functions $f_{agg}(\cdot)$: a simple averaging \citep[AVG, ][]{hosoya2018}, a product-of-experts \citep[PoE, ][]{wu_multimodal_2018}, a mixture-of-experts \citep[MoE, ][]{shi_variational_2019}, and a mixture-of-products-of-experts \citep[MoPoE, ][]{sutter2021}. Appendix \ref{app:subsec:Implementation} provides the implementation details. 

\paragraph{Setting}
\label{par:latent_eval_settings}
We train all VAEs on the same training set under identical conditions. Each VAE is trained by optimizing its respective objective, $\mathcal{E}(\Vec{X})$, as outlined in \cref{sec:methods}.
Subsequently, we assess the quality of the learned latent representations by training non-linear classifiers in a supervised setting. More specifically, we use latent representations 
of the training set to independently train binary random forest classifiers for each label and method. We assess the performance of these classifiers on the latent representations of the test set using AUROC scores.
For all the methods, we compare the quality of the unimodal representations $\vec{z}_f$ and $\vec{z}_l$. Additionally, for the aggregation-based VAEs, we evaluate the joint representation $\vec{z}_j$.


\paragraph{Results}

\begin{table*}[t]
\vspace{-1.5pt}
\caption{Evaluation of the VAE's frontal $\vec{z}_f$, lateral $\vec{z}_l$, and joint $\vec{z}_j$ representations on a subset of labels. The average AUROC [\%] and standard deviation over three seeds are reported. Full results in Appendix~\ref{app:subsec:EXP_vae}.}
\label{tab:main_table}
\vspace{0.0pt}
\small
\centering
\begin{tabular}{llcccccc}
\toprule
 &  & All Labels & No Finding & Cardiomegaly & Edema & Lung Lesion & Consolidation \\
\midrule
 & $\vec{z}_f$ & \small 68.7 \scriptsize $\pm$ 9.0 & \small 76.6 \scriptsize $\pm$ 0.3 & \small 76.3 \scriptsize $\pm$ 0.4 & \small 83.0 \scriptsize $\pm$ 0.3 & \small 61.3 \scriptsize $\pm$ 0.4 & \small 62.4 \scriptsize $\pm$ 0.4 \\
independent & $\vec{z}_l$ & \small 67.2 \scriptsize $\pm$ 7.6 & \small 73.9 \scriptsize $\pm$ 0.3 & \small 70.8 \scriptsize $\pm$ 0.9 & \small 75.4 \scriptsize $\pm$ 0.9 & \small 58.9 \scriptsize $\pm$ 0.2 & \small 64.4 \scriptsize $\pm$ 1.4 \\
 & $\vec{z}_j$ & - & - & - & - & - & - \\
\midrule
 & $\vec{z}_f$ & \small 71.0 \scriptsize $\pm$ 8.6 & \small 77.8 \scriptsize $\pm$ 0.0 & \small 78.5 \scriptsize $\pm$ 0.2 & \small 84.6 \scriptsize $\pm$ 0.3 & \small 61.8 \scriptsize $\pm$ 0.2 & \small 66.0 \scriptsize $\pm$ 0.8 \\
AVG & $\vec{z}_l$ & \small 68.7 \scriptsize $\pm$ 8.1 & \small 74.8 \scriptsize $\pm$ 0.2 & \small 73.7 \scriptsize $\pm$ 0.1 & \small 78.0 \scriptsize $\pm$ 0.3 & \small 59.0 \scriptsize $\pm$ 0.2 & \small 65.4 \scriptsize $\pm$ 1.5 \\
 & $\vec{z}_j$ & \small 69.4 \scriptsize $\pm$ 8.4 & \small 76.9 \scriptsize $\pm$ 0.4 & \small 75.2 \scriptsize $\pm$ 0.4 & \small 81.6 \scriptsize $\pm$ 0.2 & \small 61.0 \scriptsize $\pm$ 0.1 & \small 65.4 \scriptsize $\pm$ 0.8 \\
\midrule
 & $\vec{z}_f$ & \small 69.4 \scriptsize $\pm$ 8.8 & \small 77.1 \scriptsize $\pm$ 0.2 & \small 76.5 \scriptsize $\pm$ 0.6 & \small 82.4 \scriptsize $\pm$ 0.6 & \small 60.6 \scriptsize $\pm$ 0.9 & \small 62.9 \scriptsize $\pm$ 0.6 \\
MoE & $\vec{z}_l$ & \small 68.4 \scriptsize $\pm$ 8.4 & \small 75.9 \scriptsize $\pm$ 0.2 & \small 73.3 \scriptsize $\pm$ 0.2 & \small 78.0 \scriptsize $\pm$ 0.5 & \small 58.6 \scriptsize $\pm$ 0.8 & \small 64.9 \scriptsize $\pm$ 0.9 \\
 & $\vec{z}_j$ & \small 68.2 \scriptsize $\pm$ 8.2 & \small 75.8 \scriptsize $\pm$ 0.3 & \small 73.9 \scriptsize $\pm$ 0.7 & \small 79.7 \scriptsize $\pm$ 0.6 & \small 59.1 \scriptsize $\pm$ 0.5 & \small 65.1 \scriptsize $\pm$ 1.1 \\
\midrule
 & $\vec{z}_f$ & \small 70.2 \scriptsize $\pm$ 8.8 & \small 77.4 \scriptsize $\pm$ 0.1 & \small 77.1 \scriptsize $\pm$ 0.1 & \small 83.1 \scriptsize $\pm$ 0.6 & \small 60.7 \scriptsize $\pm$ 0.8 & \small 63.9 \scriptsize $\pm$ 0.3 \\
MoPoE & $\vec{z}_l$ & \small 70.3 \scriptsize $\pm$ 8.6 & \small 77.1 \scriptsize $\pm$ 0.1 & \small 75.5 \scriptsize $\pm$ 0.1 & \small 81.1 \scriptsize $\pm$ 0.8 & \small 60.8 \scriptsize $\pm$ 0.3 & \small 65.8 \scriptsize $\pm$ 0.8 \\
 & $\vec{z}_j$ & \small 70.0 \scriptsize $\pm$ 8.7 & \small 77.3 \scriptsize $\pm$ 0.1 & \small 76.4 \scriptsize $\pm$ 0.2 & \small 82.3 \scriptsize $\pm$ 0.6 & \small 60.4 \scriptsize $\pm$ 0.9 & \small 65.2 \scriptsize $\pm$ 0.1 \\
\midrule
 & $\vec{z}_f$ & \small 71.3 \scriptsize $\pm$ 8.4 & \small 77.2 \scriptsize $\pm$ 0.2 & \small 78.5 \scriptsize $\pm$ 0.3 & \small 84.5 \scriptsize $\pm$ 0.3 & \small 63.4 \scriptsize $\pm$ 0.4 & \small 66.7 \scriptsize $\pm$ 0.8 \\
PoE & $\vec{z}_l$ & \small 69.4 \scriptsize $\pm$ 8.0 & \small 74.6 \scriptsize $\pm$ 0.1 & \small 74.8 \scriptsize $\pm$ 0.1 & \small 79.1 \scriptsize $\pm$ 0.1 & \small 59.3 \scriptsize $\pm$ 0.3 & \small 66.7 \scriptsize $\pm$ 0.9 \\
 & $\vec{z}_j$ & \small 70.3 \scriptsize $\pm$ 8.9 & \small 77.5 \scriptsize $\pm$ 0.1 & \small 76.8 \scriptsize $\pm$ 0.2 & \small 83.4 \scriptsize $\pm$ 0.3 & \small 60.4 \scriptsize $\pm$ 0.7 & \small 66.2 \scriptsize $\pm$ 0.4 \\
\midrule
 & $\vec{z}_f$ & \small \textbf{73.3 }\scriptsize $\pm$ 8.9 & \small \textbf{79.1} \scriptsize $\pm$ 0.1 & \small \textbf{80.5} \scriptsize $\pm$ 0.1 & \small \textbf{86.3} \scriptsize $\pm$ 0.1 & \small \textbf{64.1} \scriptsize $\pm$ 0.2 & \small 69.1 \scriptsize $\pm$ 0.6 \\
MMVM & $\vec{z}_l$ & \small 73.0 \scriptsize $\pm$ 8.5 & \small 78.3 \scriptsize $\pm$ 0.1 & \small 78.7 \scriptsize $\pm$ 0.0 & \small 84.3 \scriptsize $\pm$ 0.3 & \small 63.0 \scriptsize $\pm$ 0.7 & \small \textbf{70.2} \scriptsize $\pm$ 0.8 \\
 & $\vec{z}_j$ & - & - & - & - & - & - \\
\bottomrule
\end{tabular}
\end{table*}

The MMVM method consistently outperforms the baselines by learning representations that lead to superior classification performance. \Cref{tab:main_table} presents the AUROC of the classifiers trained on the frontal, lateral, and joint representations ($\vec{z}_f$, $\vec{z}_l$, $\vec{z}_j$ respectively). Results suggest that $\vec{z}_f$ is more predictive than $\vec{z}_l$ for most labels. Nevertheless, MMVM's performance on the less predictive lateral modality $\vec{z}_l$ surpasses that of the other VAEs on the more predictive frontal modality $\vec{z}_f$ for labels such as \textit{Cardiomegaly} and \textit{No Finding}. This shows the MMVM's ability to soft-share information between modality-specific latent representations during training, thereby enhancing the representation of each modality.
For aggregation-based methods, the classification performance based on $\vec{z}_j$ does not outperform that of their best unimodal representation, despite having access to more information. This highlights the difficulty of preserving modality-specific details in aggregation-based methods.
In contrast, MMVM unimodal representations benefit from the complementary modality during training while successfully preserving modality-specific information, achieving superior classification performance.
Extensive results can be found in Appendix \ref{app:subsec:EXP_vae}.

\begin{figure}[htbp]
\hspace{6.5mm}
\floatconts
    {fig:main_legend}
    {}
    {\includegraphics[width=0.89\linewidth]{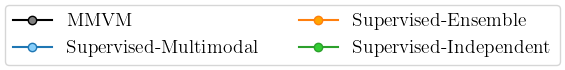}}
\vspace{-1.2cm}
\end{figure}

\begin{figure}[htbp]
\captionsetup{skip=-8pt} 
\floatconts
    {fig:mmvm_vs_clf}
    {\caption{Comparison of the MMVM VAE and fully supervised approaches at different levels of label availability. $|\vec{L}|$ denotes the number of labeled training samples available, where $|\vec{L}| \in \{10^3, 5\times10^3, 10^4, 2\times10^4, 4\times10^4, 6\times10^4, 8\times10^4, 10^5\} $. 
    }}
    {\includegraphics[width=1\linewidth]{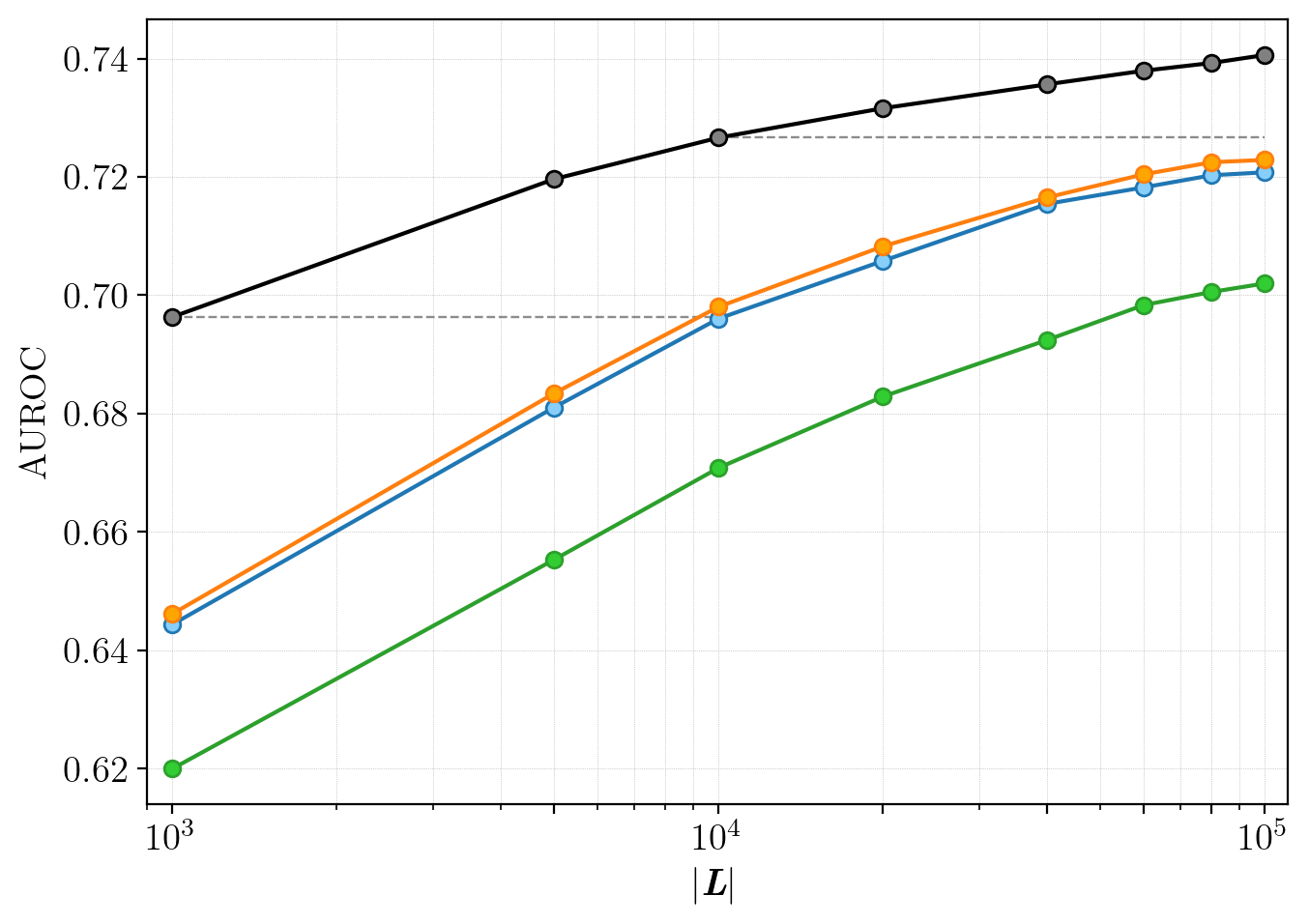}}

\end{figure}

\subsection{Impact of Label Availability}

Next, we explore the effect of different amounts of labeled data on the performance of the MMVM VAE. Specifically, we assess the effectiveness of the MMVM VAE against fully-supervised deep neural network classifiers. This comparison aims to determine whether downstream classifiers trained on the representations learned by the VAE---leveraging unlabeled data during pre-training---can match or even surpass the performance of fully supervised approaches that have to rely solely on labeled data.

\paragraph{Baselines}
We consider three fully-supervised approaches: \textit{Supervised-Independent}, which involves training independent unimodal classifiers separately; \textit{Supervised-Ensemble}, which constructs an ensemble from these independent unimodal classifiers, averaging their output scores to produce the final prediction; and \textit{Supervised-Multimodal}, which involves training a single multimodal classifier using a late fusion technique \citep{zhou2024multimodal}.
Further details are provided in Appendix \ref{app:subsec:Implementation}.

\paragraph{Setting}
We evaluate the performance of all methods under varying levels of label scarcity by progressively increasing the amount of labeled training data.
The MMVM VAE is initially trained on the full training set \emph{without label information}. Subsequently, at each evaluation step, downstream classifiers are trained on the latent representations of $\vec{L}$, where $\vec{L}$ is a subset of labeled training samples.
Fully supervised classifiers are trained directly on $\vec{L}$. Model performance is evaluated using the AUROC score on the complete test set.

\paragraph{Results} \figureref{fig:mmvm_vs_clf} presents the average AUROC across all labels for different numbers of labeled data points $|\vec{L}|$. For the \textit{Supervised-Independent} and the MMVM, the average AUROC across all labels and modalities is shown.
The MMVM VAE consistently outperforms fully supervised approaches, even when labels are available for the entire training set, demonstrating its superior ability to handle the classification task regardless of the amount of labeled data.
Among the fully supervised classifiers, the \textit{Supervised-Ensemble} slightly outperforms the \textit{Supervised-Multimodal}, with both models surpassing the independent classifier. This indicates that while complementary information exists across modalities, the \textit{Supervised-Multimodal} model fails to capture cross-modal interactions or retain modality-specific details. The MMVM shows greater stability across varying levels of label availability due to its reduced reliance on labeled data. Notably, it consistently achieves performance close to or better than the best fully supervised methods while requiring only 10\% of the labeled data. Detailed results can be found in Appendix \ref{app:subsec:EXP_label}.
\section{Conclusions}
In this work, we evaluated the Multimodal Variational Mixture-of-Experts prior (MMVM) VAE in addressing the challenges of multimodal data integration and label scarcity in medical settings, specifically using the MIMIC-CXR dataset for chest X-ray classification. The MMVM VAE demonstrated superior performance compared to baseline VAEs and fully supervised classifiers, highlighting its potential in real-world medical applications where modality-complete and fully-labeled datasets are often unavailable.
While our evaluation is promising, using a single dataset may limit the generalizability of the findings. Future research should aim to validate these results across diverse datasets and explore the MMVM VAE's application in more complex multimodal experiments, such as those involving diverse data types like electronic health records and clinical notes. Additionally, the task-independent nature of the MMVM VAE’s learned representations suggests they could be applied to a broader range of medical tasks, though this has yet to be explored.


\bibliography{jmlr-sample}
\clearpage

\appendix

\section{Dataset}
\begin{figure*}
    \centering
    \includegraphics[width=1\textwidth]{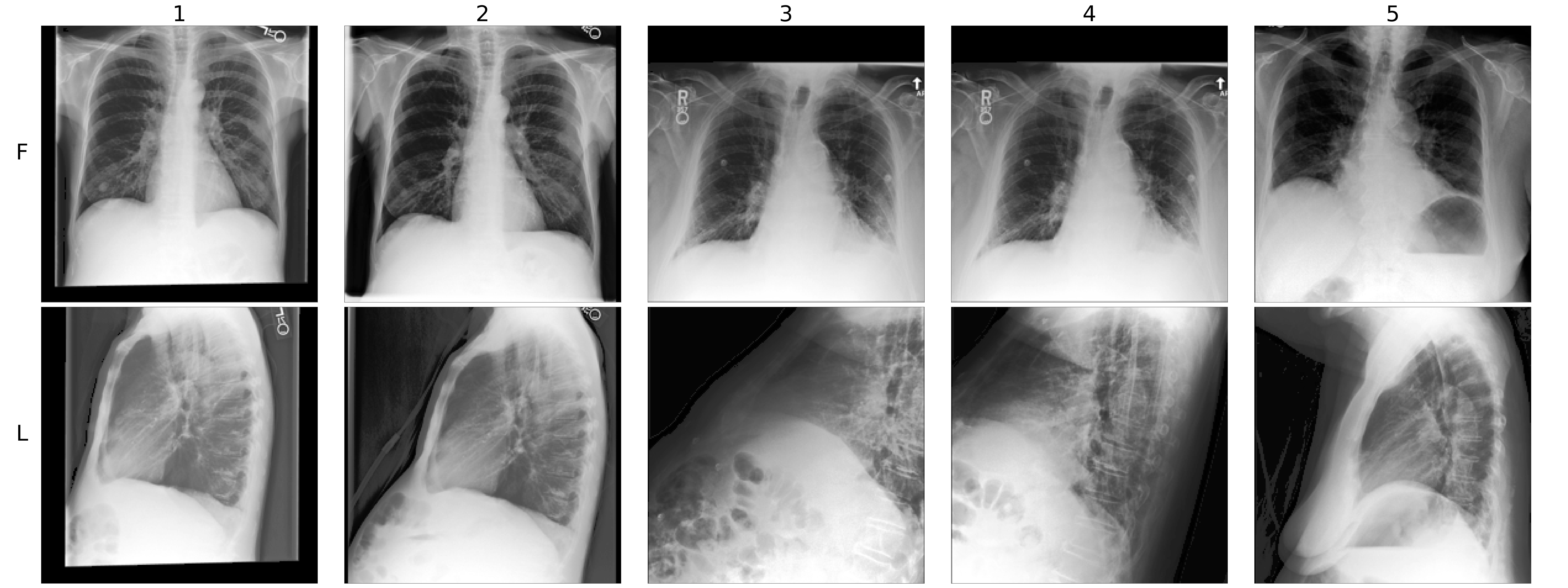}
    \caption{
    Multimodal interpretation of the MIMIC-CXR dataset. Every column is a bimodal tuple $\vec{X}$, the top row shows samples of the frontal modality $\vec{x}_f$, and the bottom row shows samples of the lateral modality $\vec{x}_l$.
    The first two tuples are linked to \textit{No Findings}, indicative of healthy conditions. Tuples three and four are labeled with \textit{Consolidation} disease. The tuple five is labeled with \textit{Atelectasis} disease. We can observe that tuples three and four share the same frontal image, but they differ due to having distinct lateral images.
    }
    \label{fig:exp_data_bimodalmimic}
\end{figure*}
\label{app:sec:Dataset}

The dataset we use in our experiment is a multimodal interpretation of the original MIMIC-CXR dataset. The original MIMIC-CXR dataset \citep{johnson2019mimic} comprises high-resolution chest X-ray images related to imaging studies. A study may include multiple chest X-ray images captured from several view positions. We categorized these views into two primary modalities: \textit{frontal} (including “AP” and “PA” views) and \textit{lateral} (including “LL” and “Lateral” views). For each study, we pair every frontal image with every lateral image in all possible combinations. Studies lacking at least one frontal and one lateral image are excluded. This approach formalizes a new dataset composed of image pairs, thus offering a bimodal interpretation of the original MIMIC-CXR dataset. More rigorously, we define a dataset $\mathbb{X} = \{ \vec{X}^{(i)} \}_{i=1}^n$ where each $\vec{X} = \{ \vec{x}_f, \vec{x}_l  \}$ is a bimodal tuple composed of one frontal image $\vec{x}_f$ and one lateral image $\vec{x}_l$ of the same study. An image may appear in multiple tuples, but we ensure that each tuple is unique by having at least one different image. Examples of these bimodal tuples are illustrated in Figure \ref{fig:exp_data_bimodalmimic}.

\begin{figure*}[htbp]
\floatconts
  {fig:subfigex2} 
  {\caption{Dataset Label Distributions. Figure 3a illustrates the original multi-class label distribution, highlighting a strong predominance of “Non-mentioned" values. Figure 3b displays the label distribution following the binarization process used in our experiments, where all non-positive labels (including “Negative,” “Non-mentioned,” and “Uncertain”) are combined into an aggregate “Negative” label.}}
  {%
    \centering
    \includegraphics[width=0.6\textwidth]{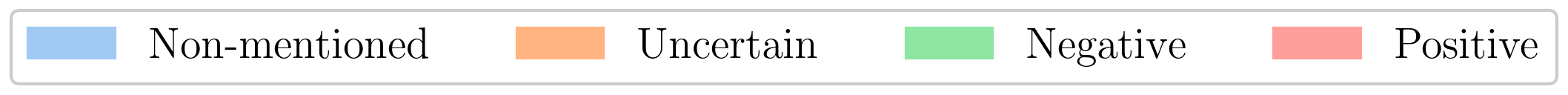} 
    \subfigure[Original Labels][c]{\label{fig:plot1}%
      \includegraphics[width=0.49\textwidth]{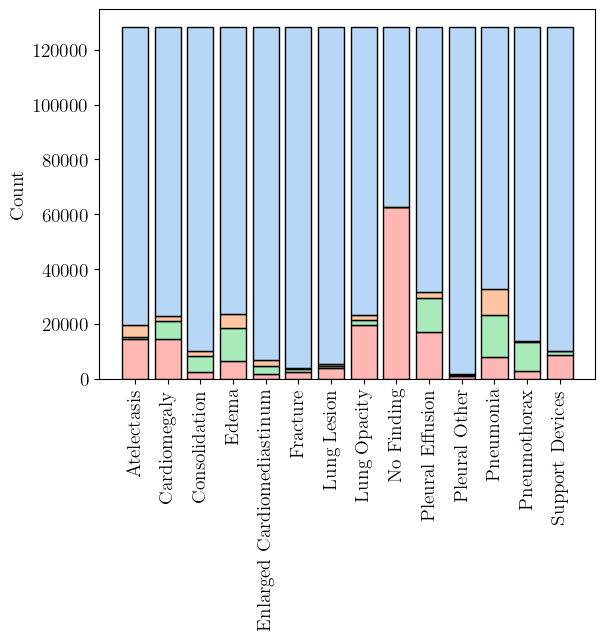}}%
    \hspace{0.01\textwidth} 
    \subfigure[Binarized Labels][c]{\label{fig:plot2}%
      \includegraphics[width=0.49\textwidth]{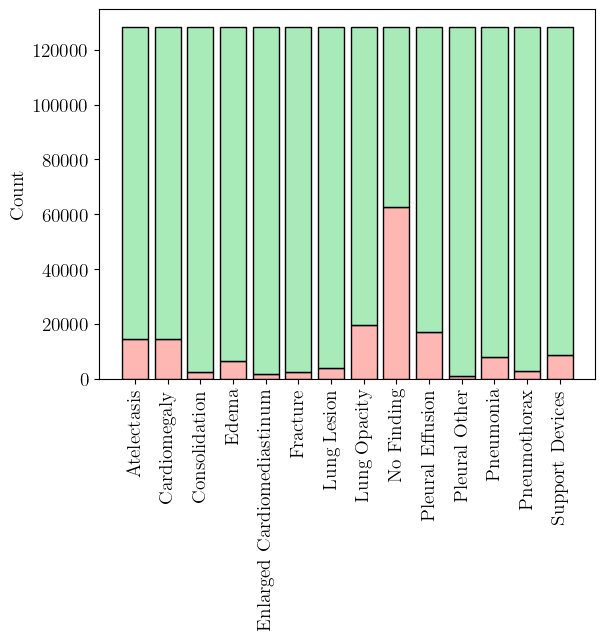}}%
  }
\label{app:fig:label_distr}
\end{figure*}

\paragraph{Preprocessing and Splitting} In our preprocessing pipeline, we apply center cropping and downscale the images to a resolution of $224\times 224$. We utilize the labels from the MIMIC-CXR-JPG dataset \citep{johnson2024mimic} which are obtained using the CheXpert tool \citep{irvin2019chexpert}. All non-positive labels (including “negative,” “non-mentioned,” or “uncertain”) are combined into an aggregate “negative” label following the approach adopted by \cite{Haque2021.07.30.21261225}. Each imaging study is connected to a subject. We split the dataset into distinct training (80\%), validation (10\%), and test (10\%) sets based on subjects, thus ensuring that the same image or study cannot be present in multiple sets. Figure \ref{app:fig:label_distr} illustrates the dataset label distribution.
\vspace{-0.2cm}

\section{Experiments}
\label{app:sec:Experiments}
\begin{table*}[t]
\caption{Evaluation of the VAEs’ frontal latent representation $\vec{z}_f$ classification performance on the
test split.
The average AUROC [\%] and standard deviation over three seeds are reported.}

\label{app:tab:frontal}
\centering
\begin{tabular}{lcccccc}
\toprule
$\vec{z}_f$ & independent & AVG & MoE & MoPoE & PoE & MMVM \\
\midrule
Atelectasis & \small73.1 \scriptsize $\pm$0.0 & \small75.2 \scriptsize $\pm$0.3 & \small73.0 \scriptsize $\pm$0.5 & \small74.2 \scriptsize $\pm$0.4 & \small75.7 \scriptsize $\pm$0.3 & \small\textbf{77.6 }\scriptsize $\pm$0.1 \\
Cardiomegaly & \small76.3 \scriptsize $\pm$0.4 & \small78.5 \scriptsize $\pm$0.2 & \small76.5 \scriptsize $\pm$0.6 & \small77.1 \scriptsize $\pm$0.1 & \small78.5 \scriptsize $\pm$0.3 & \small\textbf{80.5 }\scriptsize $\pm$0.1 \\
Consolidation & \small62.4 \scriptsize $\pm$0.4 & \small66.0 \scriptsize $\pm$0.8 & \small62.9 \scriptsize $\pm$0.6 & \small63.9 \scriptsize $\pm$0.3 & \small66.7 \scriptsize $\pm$0.8 & \small\textbf{69.1} \scriptsize $\pm$0.6 \\
Edema & \small83.0 \scriptsize $\pm$0.3 & \small84.6 \scriptsize $\pm$0.3 & \small82.4 \scriptsize $\pm$0.6 & \small83.1 \scriptsize $\pm$0.6 & \small84.5 \scriptsize $\pm$0.3 & \small\textbf{86.3} \scriptsize $\pm$0.1 \\
Enlarged Cardiomediastinum & \small59.5 \scriptsize $\pm$1.2 & \small64.9 \scriptsize $\pm$0.8 & \small61.7 \scriptsize $\pm$0.5 & \small64.1 \scriptsize $\pm$0.5 & \small66.4 \scriptsize $\pm$0.5 & \small\textbf{68.6} \scriptsize $\pm$1.1 \\
Fracture & \small56.0 \scriptsize $\pm$0.1 & \small58.4 \scriptsize $\pm$0.5 & \small57.3 \scriptsize $\pm$0.2 & \small57.4 \scriptsize $\pm$0.7 & \small\textbf{58.8} \scriptsize $\pm$0.3 & \small58.6 \scriptsize $\pm$0.8 \\
Lung Lesion & \small61.3 \scriptsize $\pm$0.4 & \small61.8 \scriptsize $\pm$0.2 & \small60.6 \scriptsize $\pm$0.9 & \small60.7 \scriptsize $\pm$0.8 & \small63.4 \scriptsize $\pm$0.4 & \small\textbf{64.1} \scriptsize $\pm$0.2 \\
Lung Opacity & \small63.8 \scriptsize $\pm$0.3 & \small65.7 \scriptsize $\pm$0.3 & \small63.5 \scriptsize $\pm$0.2 & \small64.7 \scriptsize $\pm$0.0 & \small66.3 \scriptsize $\pm$0.2 & \small\textbf{68.1} \scriptsize $\pm$0.1 \\
No Finding & \small76.6 \scriptsize $\pm$0.3 & \small77.8 \scriptsize $\pm$0.0 & \small77.1 \scriptsize $\pm$0.2 & \small77.4 \scriptsize $\pm$0.1 & \small77.2 \scriptsize $\pm$0.2 & \small\textbf{79.1} \scriptsize $\pm$0.1 \\
Pleural Effusion & \small81.2 \scriptsize $\pm$0.6 & \small82.8 \scriptsize $\pm$0.0 & \small81.6 \scriptsize $\pm$0.4 & \small82.5 \scriptsize $\pm$0.5 & \small82.8 \scriptsize $\pm$0.2 & \small\textbf{85.7} \scriptsize $\pm$0.3 \\
Pleural Other & \small67.8 \scriptsize $\pm$1.1 & \small68.9 \scriptsize $\pm$0.5 & \small67.5 \scriptsize $\pm$1.0 & \small68.3 \scriptsize $\pm$1.2 & \small68.7 \scriptsize $\pm$1.2 & \small\textbf{70.0} \scriptsize $\pm$2.0 \\
Pneumonia & \small55.3 \scriptsize $\pm$0.5 & \small57.8 \scriptsize $\pm$0.4 & \small56.4 \scriptsize $\pm$0.4 & \small57.3 \scriptsize $\pm$0.4 & \small57.5 \scriptsize $\pm$0.4 & \small\textbf{60.0} \scriptsize $\pm$0.6 \\
Pneumothorax & \small75.3 \scriptsize $\pm$1.0 & \small78.3 \scriptsize $\pm$0.6 & \small77.7 \scriptsize $\pm$0.4 & \small78.8 \scriptsize $\pm$0.7 & \small78.5 \scriptsize $\pm$0.7 & \small\textbf{81.9} \scriptsize $\pm$0.4 \\
Support Devices & \small70.8 \scriptsize $\pm$0.3 & \small73.1 \scriptsize $\pm$0.4 & \small72.7 \scriptsize $\pm$0.7 & \small73.7 \scriptsize $\pm$0.6 & \small73.8 \scriptsize $\pm$0.1 & \small\textbf{76.6} \scriptsize $\pm$0.2 \\
\midrule
All Labels & \small68.7 \scriptsize $\pm$9.0 & \small71.0 \scriptsize $\pm$8.6 & \small69.4 \scriptsize $\pm$8.8 & \small70.2 \scriptsize $\pm$8.8 & \small71.3 \scriptsize $\pm$8.4 & \small\textbf{73.3} \scriptsize $\pm$8.9 \\
\bottomrule
\end{tabular}
\end{table*}
\vspace{-0.1cm}

\subsection{Implementation and Training}
\label{app:subsec:Implementation}
\paragraph{General}
We use the scikit-learn \citep{scikit-learn} package for the Random Forest classifiers implemented in this work. All code is written using Python 3.11, PyTorch \citep{pytorch} and Pytorch-Lightning \citep{PyTorch_Lightning_2019}. The \hyperlink{https://github.com/agostini335/mmvmvae-mimic}{code}\footnote{\scriptsize \url{https://github.com/agostini335/mmvmvae-mimic}} used to run our experiments is publicly available. We base the implementations of all the methods on the official implementations.
Hence, we base our implementations on the following repositories:
\begin{itemize}
    \item MMVM: \url{https://github.com/thomassutter/mmvmvae}
    \item AVG: \url{https://github.com/HaruoHosoya/gvae}
    \item PoE: \url{https://github.com/mhw32/multimodal-vae-public}
    \item MoE: \url{https://github.com/iffsid/mmvae}
    \item MoPoE: \url{https://github.com/thomassutter/MoPoE}
\end{itemize}
\vspace{-0.1cm}
\paragraph{VAEs} All the VAEs used in this work are built using the the same architectural design and trained under identical conditions. We use ResNet-based encoders and decoders \citep{he_deep_2016}. The encoder and decoder consist of 2D convolution layers. The design of the encoders and decoders is uniform for both frontal and lateral modalities. We use an Adam optimizer \citep{kingma_adam_2014} with a learning rate of $0.00005$ and a batch size of $32$. We train all methods for $240$ epochs and $3$ seeds. We use NVIDIA A100-SXM4-40GB GPUs for all our runs. An average run, evaluating one method on one seed, takes approximately $45$ hours. To train all methods evaluated in this experiment, we had to train $3 \times 6 = 18$ different models: $3$ seeds and $6$ methods. Hence, the total GPU compute time used to generate the VAE results is around $45 \times 18 = 810$ hours. We also had to use GPU time in the development process, which we did not measure.

\paragraph{Random Forest Downstream Task Classifiers}

In our experiments, we focus on the diagnostic label prediction task to evaluate the quality of the latent representations learned by the VAEs. For this purpose, we use Random Forest (RF) classifiers to perform the downstream classification task. Specifically, we train independent RF binary classifiers for each method and each label on the inferred representations of the training set and evaluate them on the representations of the test set. The RF classifiers are configured with 5,000 estimators and a maximum depth of 30. In Experiment \ref{sec:latent_representations_eval}, we train each RF classifier using 20,000 latent representations.

\paragraph{Fully-Supervised Deep Neural Network Classifiers}
We evaluate three fully-supervised approaches: \textit{Supervised-Independent}, where independent unimodal classifiers are trained separately; \textit{Supervised-Ensemble}, which averages the output scores of these independent classifiers to generate the final prediction; and \textit{Supervised-Multimodal}, where a multimodal classifier is trained using a late fusion technique. In the \textit{Supervised-Multimodal} approach, modalities are fused during training by averaging their modality-specific network final layer outputs into a shared last representation, allowing the model to learn cross-modal interactions. This differs from the \textit{Supervised-Ensemble}, where the output scores from independent classifiers are combined only after training.
All the fully supervised approaches share the same architectural design. For both the frontal and lateral modalities, the network architecture is derived from the VAE encoder used in this work. To adapt the encoder for the classification task, a linear layer with 14 neurons corresponding to the number of labels in the MIMIC-CXR dataset is added. We use the Adam optimizer with a learning rate of $0.0001$ and a batch size of $256$. Each model is trained until the best AUROC on the validation set is achieved.

\vspace{-0.37cm}

\subsection{Additional Results}
In the following section, we provide additional results and details for the two experiments conducted in this work: \textit{Comparison of Multimodal VAEs} and \textit{Impact of Label Availability}.

\subsubsection{Comparison of Multimodal VAEs}
\label{app:subsec:EXP_vae}
In Tables \ref{app:tab:frontal}, \ref{app:tab:lateral}, \ref{app:tab:joint}, we report the classification performance obtained using the frontal, lateral, and joint latent representations of the different VAEs. Each table reports the average AUROC scores along with the standard deviation across three seeds for all labels.

\vspace{-0.37cm}

\subsubsection{Impact of Label Availability}
\label{app:subsec:EXP_label}
\begin{figure*}[htbp]

\floatconts
  {fig:subfigex2} 
  {\caption{The effect of label availability on the MMVM and fully-supervised approaches. The X-axis is presented on a logarithmic scale for clearer visualization. The average AUROC and standard deviation across four seeds are reported for each label. $|\vec{L}|$ denotes the number of labeled training
samples available, where $|\vec{L}| \in \{10^3, 5\times10^3, 10^4, 2\times10^4, 4\times10^4, 6\times10^4, 8\times10^4, 10^5\} $.}}
  {%
    \subfigure[MMVM][c]{\label{fig:plot1}%
      \includegraphics[width=0.47\textwidth]{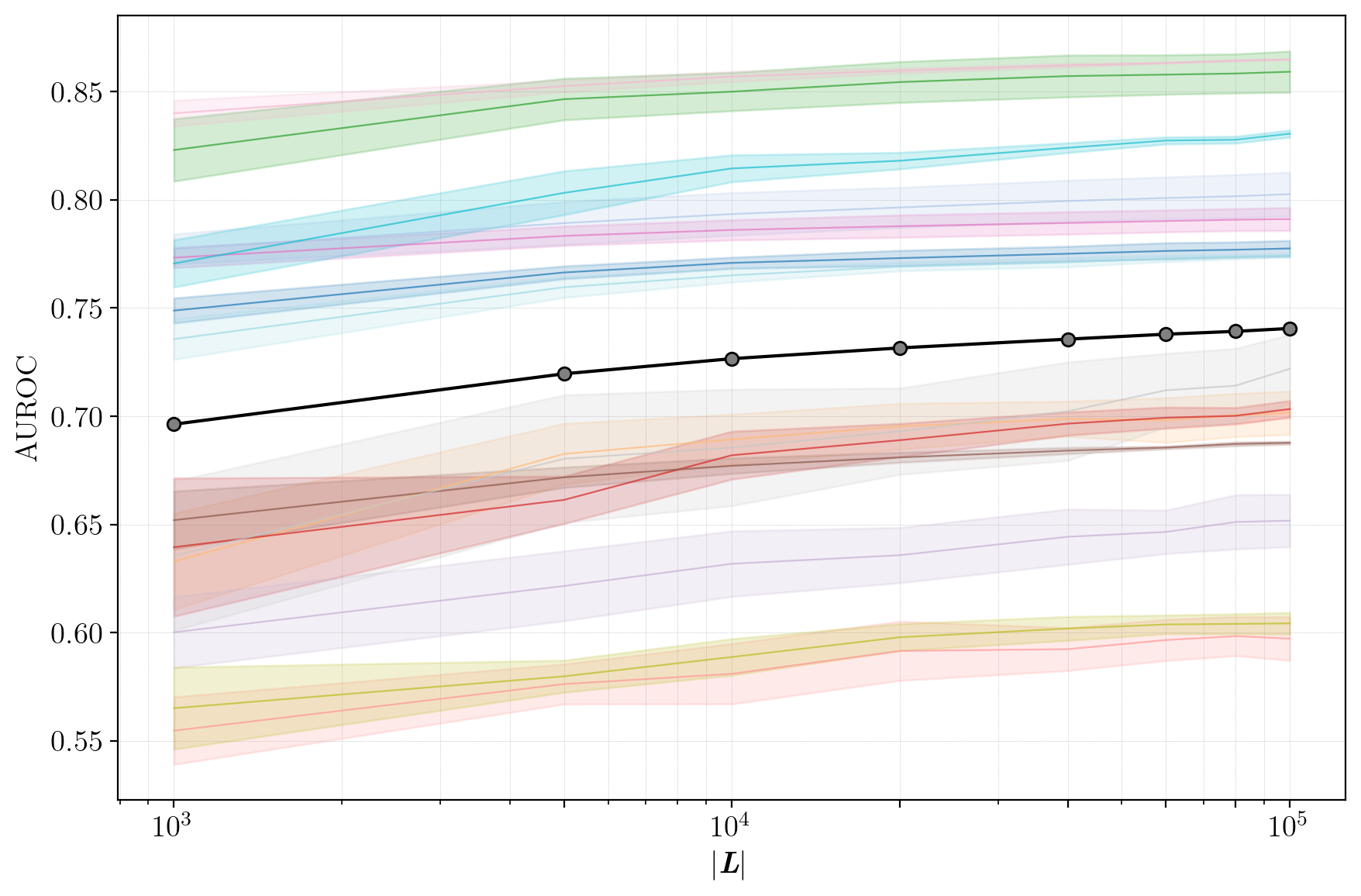}}%
    \qquad
    \subfigure[Supervised-Ensemble][c]{\label{fig:plot2}%
      \includegraphics[width=0.47\textwidth]{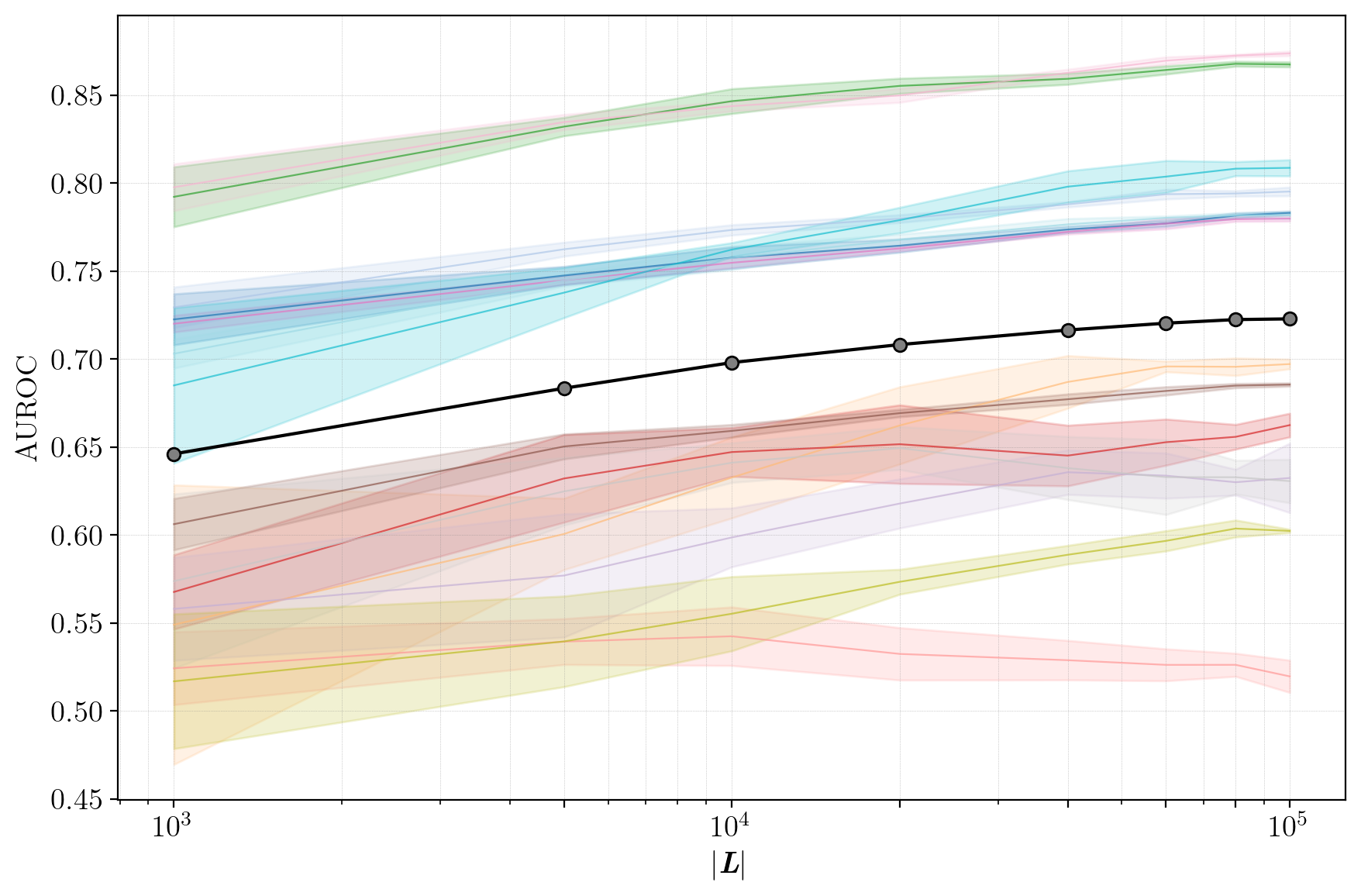}}
    \vspace{1cm}  
    \subfigure[Supervised-Multimodal][c]
    {\label{fig:plot3}%
      \includegraphics[width=0.47\textwidth]{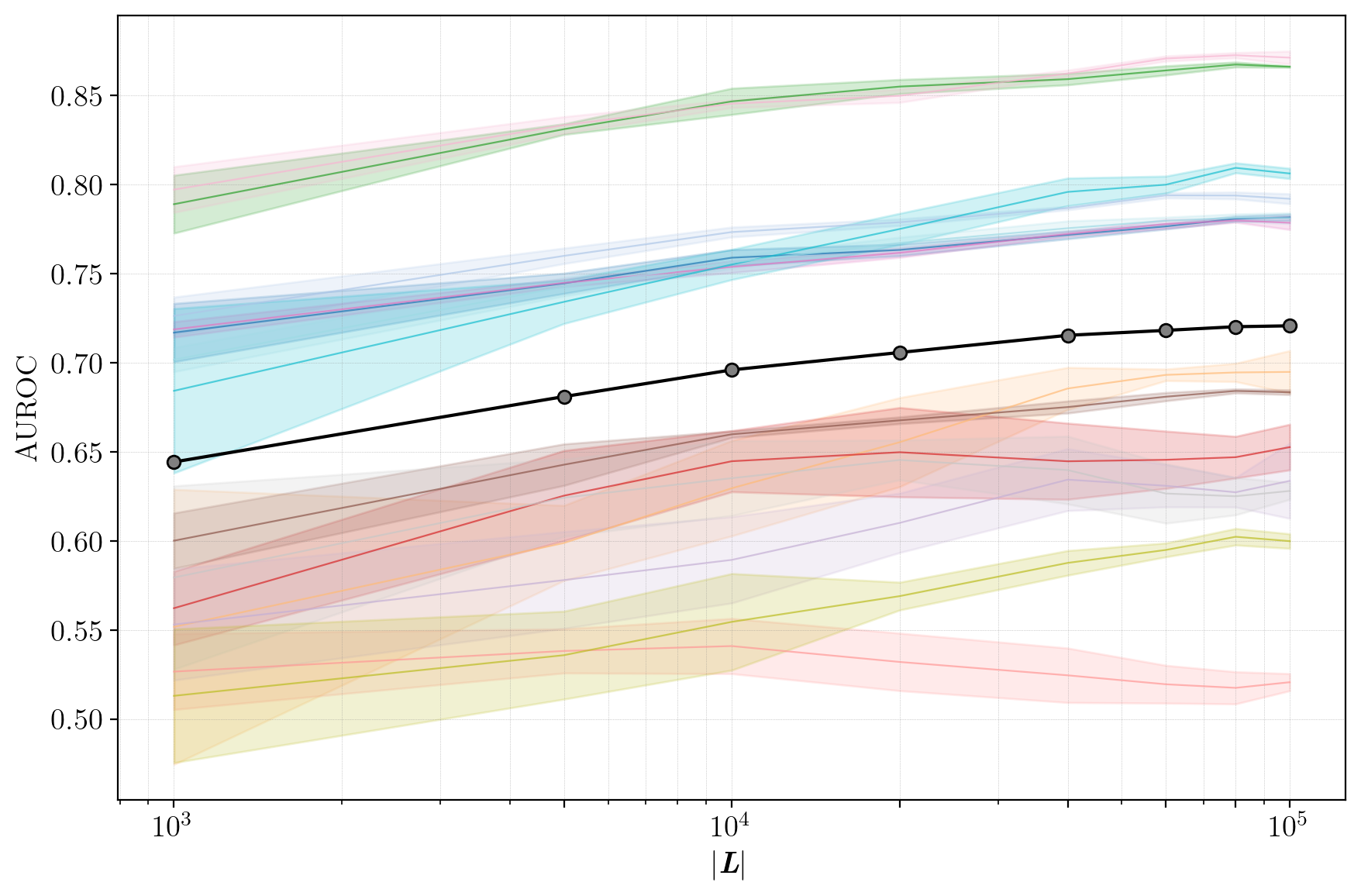}}%
    \qquad
    \subfigure[Supervised-Independent][c]{\label{fig:plot4}%
      \includegraphics[width=0.47\textwidth]{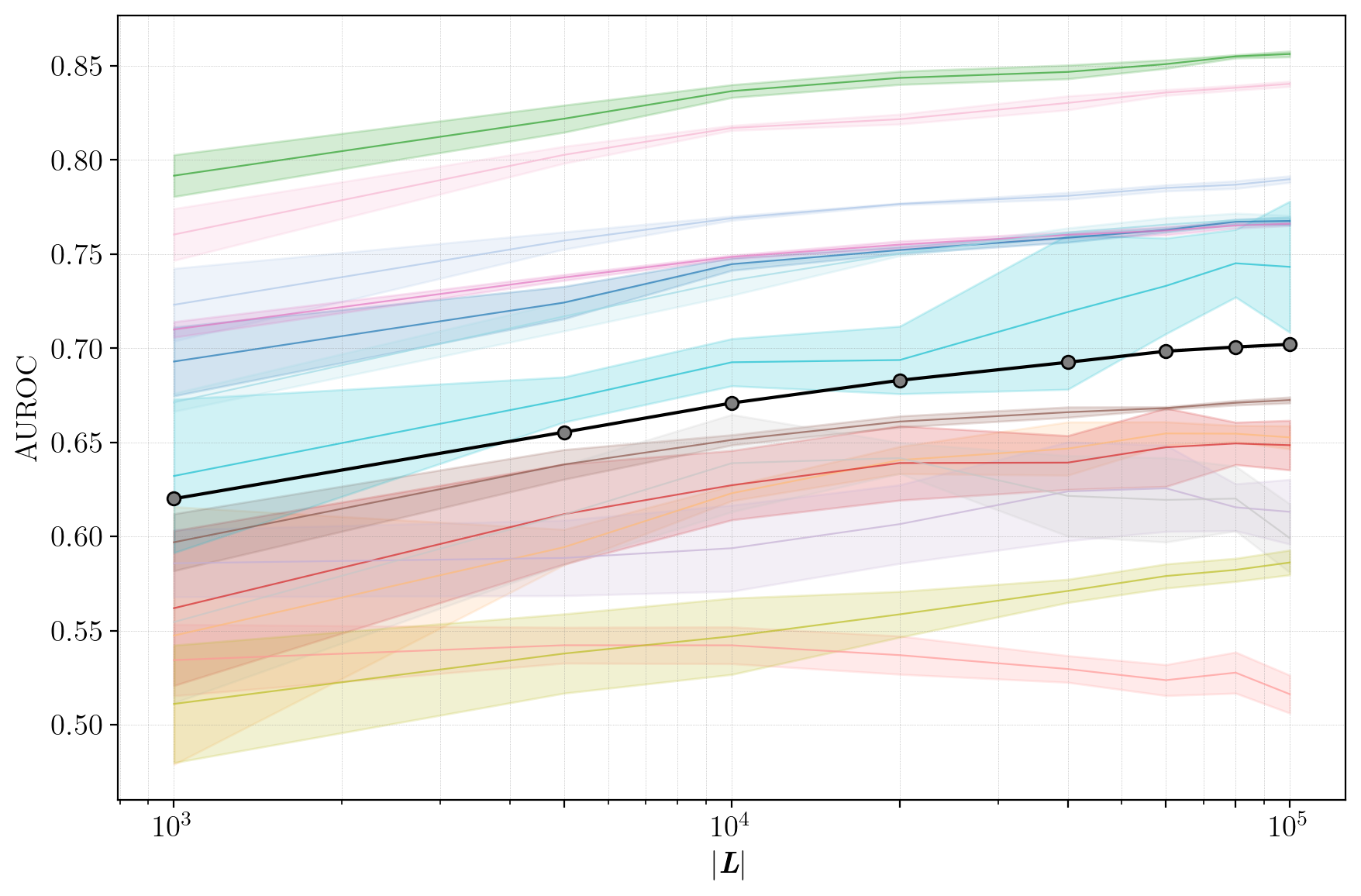}}
    \vspace{0.5cm} 
    \centering
    \includegraphics[width=0.9\textwidth]{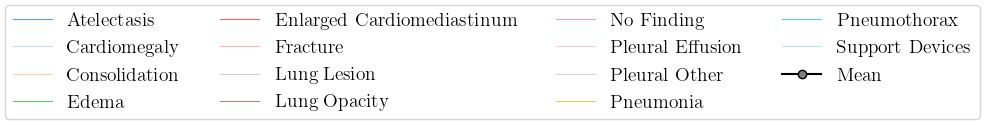} 
  }
  \label{app:fig:labels_exp}
\end{figure*}

Figure \ref{app:fig:labels_exp} shows the AUROC for each label achieved by the MMVM and fully-supervised classifiers at varying levels of label availability. For both the \textit{Supervised-Independent} classifier and the MMVM, the average AUROC across frontal and lateral modalities is displayed. The figure highlights the superior performance of the MMVM over the fully-supervised approach, as well as its stability across different amounts of labeled data for all fourteen labels. Additional information regarding the dataset label distribution is provided in Figure \ref{app:fig:label_distr}.

\vspace{-0.37cm}
\section{Generative Results}

\begin{figure*}[htbp]
\centering
\floatconts
  {fig:subfigex2} 
  {\caption{Qualitative Results on the lateral-to-frontal image generation task. The
first 3 rows of every subplot show the input images and the bottom 3 rows the conditional generations.
Different from the training, we provide only the lateral image to every model and based on the latent representation $\vec{z}_l$ of that image, we generate a frontal image sample. We see that the aggregated VAE (Figure 5b) is not able
to conditionally generate visually pleasing samples compared to the independent VAEs (Figure 5a)
and the MMVM VAE (Figure 5c). The samples generated by the independent VAEs appear to be more detailed.}}
  {%
    \subfigure[Independent VAEs][c]{\label{fig:plot1}%
      \includegraphics[width=0.32\textwidth]{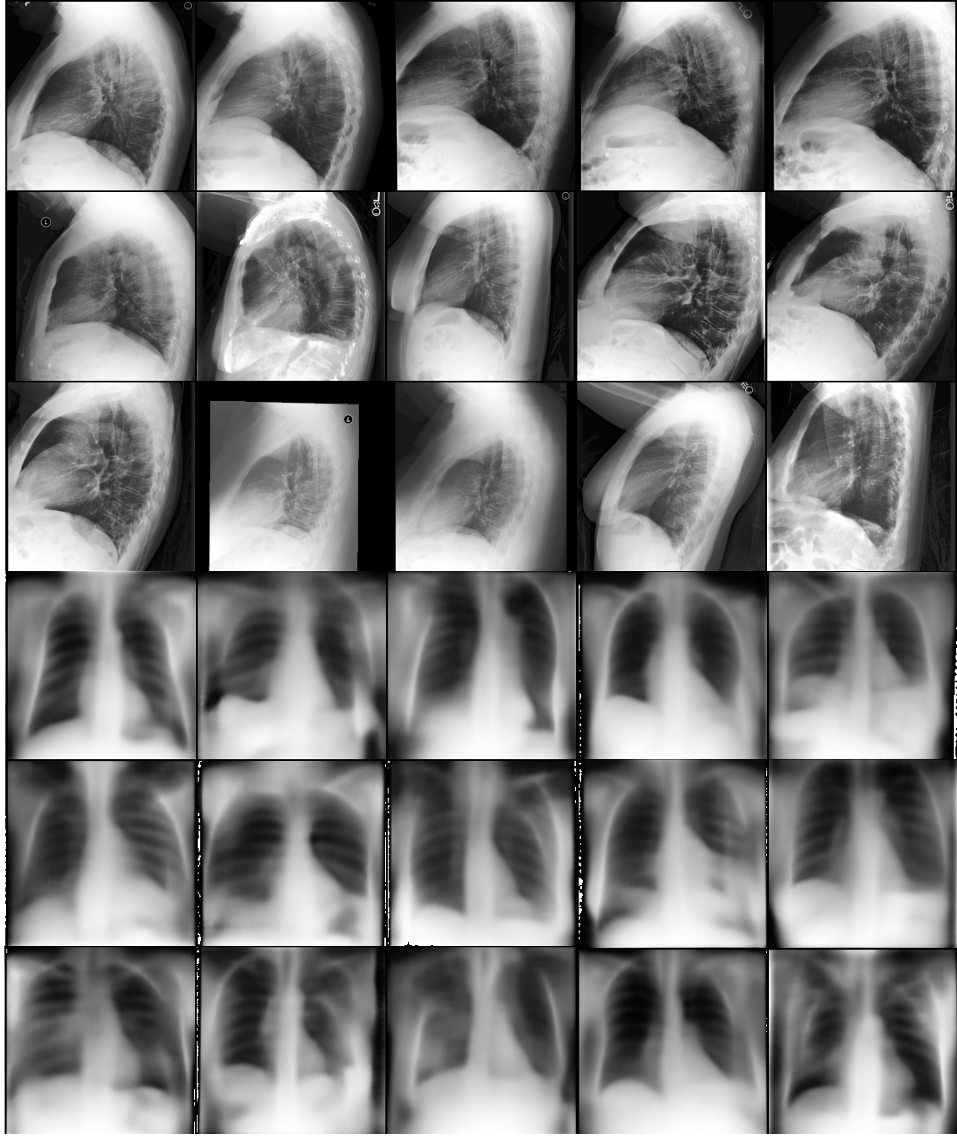}}%
      \hspace{0.01\textwidth} 
    \subfigure[PoE VAE][c]{\label{fig:plot2}%
      \includegraphics[width=0.32\textwidth]{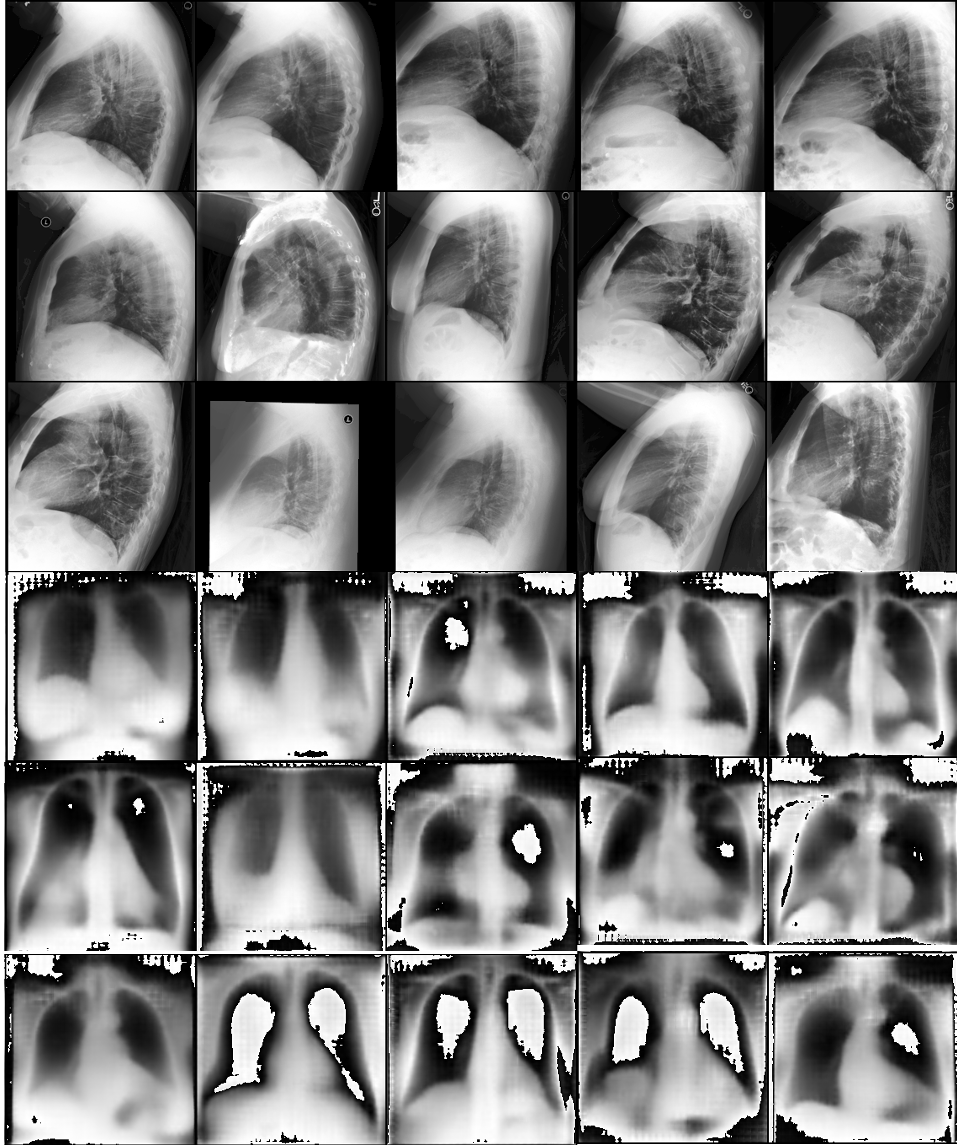}}%
      \hspace{0.01\textwidth} 
    \subfigure[MMVM VAE][c]{\label{fig:plot3}%
      \includegraphics[width=0.32\textwidth]{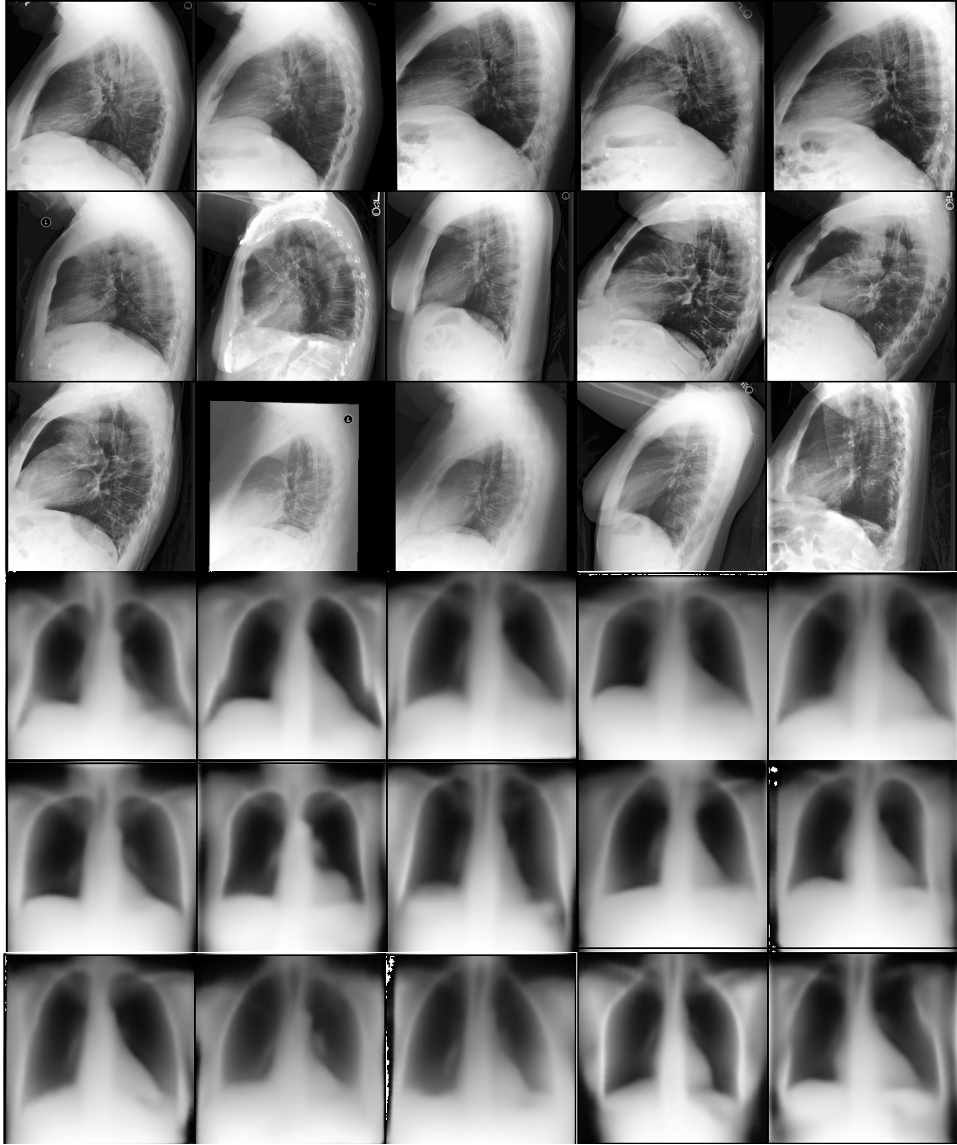}}%
  }
\label{app:fig:conditional_gen}
\end{figure*}

The MMVM VAE allows conditional generation, which can be particularly valuable in medical contexts for generating missing modalities. In Figure \ref{app:fig:conditional_gen}, we qualitatively evaluate the generation of frontal images based on given lateral images. We compare the MMVM VAE results with those produced by independent VAEs and an aggregation-based multimodal VAE using the PoE aggregation function.

\begin{table*}[htbp!]

\caption{Evaluation of the VAEs’ joint latent representation $\vec{z}_l$ classification performance on the
test split.
The average AUROC [\%] and standard deviation over three seeds are reported.}

\label{app:tab:lateral}
\centering

\begin{tabular}{lcccccc}
\toprule
$\vec{z}_l$  & independent & AVG & MoE & MoPoE & PoE & MMVM \\
\midrule
Atelectasis & \small70.7 \scriptsize $\pm$0.3 & \small73.5 \scriptsize $\pm$0.4 & \small72.8 \scriptsize $\pm$0.1 & \small74.7 \scriptsize $\pm$0.2 & \small73.7 \scriptsize $\pm$0.2 & \small\textbf{77.0} \scriptsize $\pm$0.2 \\
Cardiomegaly & \small70.8 \scriptsize $\pm$0.9 & \small73.7 \scriptsize $\pm$0.1 & \small73.3 \scriptsize $\pm$0.2 & \small75.5 \scriptsize $\pm$0.1 & \small74.8 \scriptsize $\pm$0.1 & \small\textbf{78.7} \scriptsize $\pm$0.0 \\
Consolidation & \small64.4 \scriptsize $\pm$1.4 & \small65.4 \scriptsize $\pm$1.5 & \small64.9 \scriptsize $\pm$0.9 & \small65.8 \scriptsize $\pm$0.8 & \small66.7 \scriptsize $\pm$0.9 & \small\textbf{70.2} \scriptsize $\pm$0.8 \\
Edema & \small75.4 \scriptsize $\pm$0.9 & \small78.0 \scriptsize $\pm$0.3 & \small78.0 \scriptsize $\pm$0.5 & \small81.1 \scriptsize $\pm$0.8 & \small79.1 \scriptsize $\pm$0.1 & \small\textbf{84.3} \scriptsize $\pm$0.3 \\
Enlarged Cardiomediastinum & \small60.1 \scriptsize $\pm$0.7 & \small62.0 \scriptsize $\pm$1.0 & \small60.5 \scriptsize $\pm$0.5 & \small64.2 \scriptsize $\pm$0.9 & \small63.5 \scriptsize $\pm$0.8 & \small\textbf{69.0} \scriptsize $\pm$0.7 \\
Fracture & \small57.9 \scriptsize $\pm$0.6 & \small58.3 \scriptsize $\pm$0.7 & \small56.8 \scriptsize $\pm$0.8 & \small58.6 \scriptsize $\pm$0.8 & \small59.0 \scriptsize $\pm$0.5 & \small\textbf{60.9} \scriptsize $\pm$0.3 \\
Lung Lesion & \small58.9 \scriptsize $\pm$0.2 & \small59.0 \scriptsize $\pm$0.2 & \small58.6 \scriptsize $\pm$0.8 & \small60.8 \scriptsize $\pm$0.3 & \small59.3 \scriptsize $\pm$0.3 & \small\textbf{63.0} \scriptsize $\pm$0.7 \\
Lung Opacity & \small61.9 \scriptsize $\pm$0.5 & \small63.4 \scriptsize $\pm$0.4 & \small63.9 \scriptsize $\pm$0.1 & \small65.4 \scriptsize $\pm$0.4 & \small64.1 \scriptsize $\pm$0.4 & \small\textbf{68.1} \scriptsize $\pm$0.2 \\
No Finding & \small73.9 \scriptsize $\pm$0.3 & \small74.8 \scriptsize $\pm$0.2 & \small75.9 \scriptsize $\pm$0.2 & \small77.1 \scriptsize $\pm$0.1 & \small74.6 \scriptsize $\pm$0.1 & \small\textbf{78.}3 \scriptsize $\pm$0.1 \\
Pleural Effusion & \small80.2 \scriptsize $\pm$0.2 & \small82.0 \scriptsize $\pm$0.1 & \small82.0 \scriptsize $\pm$0.3 & \small84.3 \scriptsize $\pm$0.2 & \small82.1 \scriptsize $\pm$0.1 & \small\textbf{85.7} \scriptsize $\pm$0.1 \\
Pleural Other & \small62.8 \scriptsize $\pm$1.3 & \small64.3 \scriptsize $\pm$0.7 & \small62.7 \scriptsize $\pm$1.7 & \small63.6 \scriptsize $\pm$1.0 & \small63.9 \scriptsize $\pm$1.0 & \small\textbf{68.5} \scriptsize $\pm$1.9 \\
Pneumonia & \small56.4 \scriptsize $\pm$0.5 & \small56.9 \scriptsize $\pm$0.3 & \small57.5 \scriptsize $\pm$1.2 & \small58.3 \scriptsize $\pm$0.5 & \small58.2 \scriptsize $\pm$0.1 & \small\textbf{59.0} \scriptsize $\pm$0.2 \\
Pneumothorax & \small75.6 \scriptsize $\pm$0.5 & \small77.8 \scriptsize $\pm$0.3 & \small76.9 \scriptsize $\pm$0.7 & \small79.2 \scriptsize $\pm$0.6 & \small78.6 \scriptsize $\pm$0.2 & \small\textbf{81.7} \scriptsize $\pm$0.3 \\
Support Devices & \small71.9 \scriptsize $\pm$0.6 & \small72.9 \scriptsize $\pm$0.5 & \small73.6 \scriptsize $\pm$0.7 & \small75.9 \scriptsize $\pm$0.4 & \small74.7 \scriptsize $\pm$0.5 & \small\textbf{77.1} \scriptsize $\pm$0.3 \\
\midrule
All Labels & \small67.2 \scriptsize $\pm$7.6 & \small68.7 \scriptsize $\pm$8.1 & \small68.4 \scriptsize $\pm$8.4 & \small70.3 \scriptsize $\pm$8.6 & \small69.4 \scriptsize $\pm$8.0 & \small\textbf{73.0} \scriptsize $\pm$8.5 \\
\bottomrule
\end{tabular}

\end{table*}

\begin{table*}[htbp!]

\caption{Evaluation of the VAEs’ joint latent representation $\vec{z}_j$ classification performance on the
test split.
The average AUROC [\%] and standard deviation over three seeds are reported.}
\label{app:tab:joint}
\centering

\begin{tabular}{lcccc}
\toprule
$\vec{z}_j$  & AVG & MoE & MoPoE & PoE \\
\midrule
Atelectasis & \small73.0 \scriptsize $\pm$0.3 & \small71.9 \scriptsize $\pm$0.1 & \small74.0 \scriptsize $\pm$0.0 & \small\textbf{74.8} \scriptsize $\pm$0.2 \\
Cardiomegaly & \small75.2 \scriptsize $\pm$0.4 & \small73.9 \scriptsize $\pm$0.7 & \small76.4 \scriptsize $\pm$0.2 & \small\textbf{76.8} \scriptsize $\pm$0.2 \\
Consolidation & \small65.4 \scriptsize $\pm$0.8 & \small65.1 \scriptsize $\pm$1.1 & \small65.2 \scriptsize $\pm$0.1 & \small\textbf{66.2} \scriptsize $\pm$0.4 \\
Edema & \small81.6 \scriptsize $\pm$0.2 & \small79.7 \scriptsize $\pm$0.6 & \small82.3 \scriptsize $\pm$0.6 & \small\textbf{83.4} \scriptsize $\pm$0.3 \\
Enlarged Cardiomediastinum & \small60.3 \scriptsize $\pm$0.5 & \small60.1 \scriptsize $\pm$0.9 & \small63.4 \scriptsize $\pm$0.6 & \small\textbf{60.8} \scriptsize $\pm$0.2 \\
Fracture & \small\textbf{59.7} \scriptsize $\pm$0.1 & \small57.1 \scriptsize $\pm$0.6 & \small57.8 \scriptsize $\pm$0.4 & \small59.2 \scriptsize $\pm$0.6 \\
Lung Lesion & \small\textbf{61.0} \scriptsize $\pm$0.1 & \small59.1 \scriptsize $\pm$0.5 & \small60.4 \scriptsize $\pm$0.9 & \small60.4 \scriptsize $\pm$0.7 \\
Lung Opacity & \small63.5 \scriptsize $\pm$0.2 & \small63.1 \scriptsize $\pm$0.3 & \small64.6 \scriptsize $\pm$0.3 & \small\textbf{64.7} \scriptsize $\pm$0.2 \\
No Finding & \small76.9 \scriptsize $\pm$0.4 & \small75.8 \scriptsize $\pm$0.3 & \small77.3 \scriptsize $\pm$0.1 & \small\textbf{77.5} \scriptsize $\pm$0.1 \\
Pleural Effusion & \small82.3 \scriptsize $\pm$0.3 & \small80.4 \scriptsize $\pm$0.1 & \small82.7 \scriptsize $\pm$0.6 & \small\textbf{83.7} \scriptsize $\pm$0.4 \\
Pleural Other & \small65.7 \scriptsize $\pm$0.6 & \small63.4 \scriptsize $\pm$1.0 & \small65.9 \scriptsize $\pm$0.9 & \small\textbf{67.0} \scriptsize $\pm$0.4 \\
Pneumonia & \small56.6 \scriptsize $\pm$0.5 & \small56.9 \scriptsize $\pm$0.4 & \small56.7 \scriptsize $\pm$0.9 & \small\textbf{57.5} \scriptsize $\pm$0.3 \\
Pneumothorax & \small77.6 \scriptsize $\pm$0.3 & \small75.6 \scriptsize $\pm$1.0 & \small78.5 \scriptsize $\pm$0.7 & \small\textbf{79.0} \scriptsize $\pm$0.3 \\
Support Devices & \small72.4 \scriptsize $\pm$0.3 & \small72.4 \scriptsize $\pm$0.4 & \small\textbf{74.0} \scriptsize $\pm$0.4 & \small73.7 \scriptsize $\pm$0.5 \\
\midrule
All Labels & \small69.4 \scriptsize $\pm$8.4 & \small68.2 \scriptsize $\pm$8.2 & \small70.0 \scriptsize $\pm$8.7 & \small\textbf{70.3} \scriptsize $\pm$8.9 \\
\bottomrule
\end{tabular}

\end{table*}

\end{document}